# Artificial Neural Networks and Support Vector Machines for Water Demand Time Series Forecasting

Ishmael S. Msiza, Fulufhelo V. Nelwamondo and Tshilidzi Marwala

*Abstract* – Water plays a pivotal role in many physical processes, and most importantly in sustaining human life, animal life and plant life. Water supply entities therefore have the responsibility to supply clean and safe water at the rate required by the consumer. It is therefore necessary to implement mechanisms and systems that can be employed to predict both short-term and long-term water demands. The increasingly growing field of computational intelligence techniques has been proposed as an efficient tool in the modelling of dynamic phenomena. The primary objective of this paper is to compare the efficiency of two computational intelligence techniques in water demand forecasting. The techniques under comparison are the Artificial Neural Networks (ANNs) and the Support Vector Machines (SVMs). In this study it was observed that the ANNs perform better than the SVMs. This performance is measured against the generalisation ability of the two.

*Keywords*: Water Demand Forecasting, Artificial Neural Networks, Support Vector Machines, Artificial Neural Genius, Support Vector Genius, Overall Genius

## I. INTRODUCTION

The modeling of water resource variables is a very broad field that includes modeling of water quality, water demand, water reticulation networks, to mention but a few. This paper is focused on the modeling of only one water resource variable which is water demand and the study is restricted to South Africa's Gauteng Province. The Republic of South Africa has now of late been experiencing a situation whereby the demand of water is much higher than the rate at which the water is being supplied [1]. This is attributable to a number of factors such as the average annual rainfall of 497mm which is way below the world's average of 860mm [2]. However, most of the factors that contribute towards the water demand exceeding the water supply are due to human interventions. These include population growth and the economic expansion of the South African citizens, especially in the Gauteng Province. The more affluent people become; the more water they will use [3], and the more the population grows, there will be an increased demand for water. The province of Gauteng is of particular interest because of its status as the industrial powerhouse of South Africa and it houses and provides employment to almost a quarter of the South African population, some 9 million people [4].

The Gauteng Province consumes about 86% of the total water supply provided by a bulk supplier called Rand Water. With the current population growth rate of 3.13% per annum [4], the water demand in this province is definitely set to increase. Another factor that has a major influence on the demand of water is the issue of HIV/AIDS. An increase in the HIV/AIDS related deaths can have a negative effect on the population growth rate. This therefore implies that the population growth rate will not always be positive, but can at times be negative. An approach that can be employed to offset the effects of this population dynamics is to develop two models, one with the effects of HIV/AIDS neglected and another one with these effects taken into account. This will result in a reliable model because the actual water demand will be inside the envelope formed by these two extremes.

## II. LITERATURE INSPECTION

The modelling of water resource variables is a very active field of study and definitely there still is a lot of work to be done. In the initial stages, modelling of water resource variables was done using the traditional statistical models. In recent years, modern techniques have been proposed as efficient modelling tools. There is a large pool of these techniques, and hence there is always a need to investigate which technique is the most efficient for a particular application.

Gamal El-Din *et al* [5] used artificial neural networks to model wastewater treatment processes. This was a comparative study between conventional deterministic models and artificial neural networks. They observed that, in addition to the information contained in the conventional models, neural networks contained a great deal of additional information with regard to the system being modelled. Jain *et al* [6] used artificial neural networks to model the short-term water demand at the Indian Institute of Technology (IIT) in Kanpur, India. Six neural network models, five regression models and two time series models were developed and compared. All the neural network models generally displayed better performance when measured against the other models. Maier *et al* [7] conducted a study reviewing 43 research papers that employed neural networks in the prediction and forecasting of water resources variables. They observed that neural network models always work well and their use in the study of water is on the increase due to their ability to handle large amounts of non-linear, non-parametric data.

Khan and Coulibaly [8] conducted a comparative study between support vector machines, artificial neural networks and the traditional seasonal autoregressive model (SAR) in the forecasting of lake water levels. They observed that the support vector machine is generally compatible with the other two models, but when it comes to long-term forecasting, the support vector machine displays better performance. Mukherjee *et al* [9] conducted a study to predict chaotic time series using support vector machines. The performance of support vector machines stood out when compared to other approximation methods such as polynomial and rational approximation, local polynomial techniques and artificial neural networks. Other forecasting applications that employed support vector machines include the work of Mohandes *et al* in the prediction of wind speed [10]. They observed that the performance of the support vector machines is comparable to that of artificial neural networks.

All of these studies confirm that there is a need to compare the performance of various approximation techniques. The study that lead to this paper carries some element of novelty since it is the first one to carry out water demand forecasting using computational intelligence techniques in the Republic of South Africa.

## III. THEORETICAL FOUNDATION

### A. Water Scarcity

The scarcity of water in the Republic of South Africa is also soaring to new heights, especially in the Gauteng Province. In order to offset the effects of this scarcity, Rand Water has introduced the idea of supplementary water schemes. Since 1974, the water in the Vaal River has been supplemented through the inter-basin transfer of water from the Tugela River in the Kwa-Zulu Natal Province. This is what became to be known as the Tugela-Vaal Transfer Scheme [11]. Another transfer scheme takes water from the Orange River in Lesotho to supplement the Vaal dam. This is what came to be known as the Lesotho Highlands Water Project [12].

The development of supplementary water schemes is indicative of the fact that the issue of water scarcity in the Republic of South Africa is a serious one. This therefore implies that there is an urgent need for the development of tools that will assist in the effective management of water resources, and artificial neural networks have a significant role to play to that effect.

### B. Regression Approximation

Unlike using conventional software development techniques to make programs, learning methodology uses examples to synthesize these programs. The particular case where the examples are input-output pairs is called supervised learning. There are different types of learning problems and these are binary classification, multi-class classification and regression [13]. Binary classification is a problem with binary (1 or 0; true or false; LOW or HIGH) outputs. Multi-class classification is a problem with a finite number of outputs, and regression is a problem with real-valued outputs. Water demand forecasting can be regarded as a regression problem because the water time series has non-linear nature and hence the output of the predicting model has to be a real value depicting the amount of water that will be needed on a specified date.

### C. The Theory of Artificial Neural Networks in Regression

Artificial Neural Networks (ANNs) are mathematical models that can be employed in the modeling of complex systems. They can be used both for classification and regression problems. ANNs consist of three layers, namely, the input layer, the hidden layer and the output layer. The input layer represents the model inputs and the output layer represents the model outputs. The hidden layer consists of nodes that, during optimization, attempt to functionally map the model inputs to the model outputs. There are numerous ANN architectures but this study focuses on only two architectures. These are the multi-layer perceptron and the radial basis function.

*1) The multi-layer perceptron (MLP)*

Networks that have more than one layer of adaptive weights are known as multi-layer perceptrons. A multi-layer perceptron has three layers of units taking values in the range (0 to 1). Each layer is nourished with the previous layers, and hence it is also called a Jump Connection Network (JCN) [14]. MLPs can have any number of weighted connections, but networks with only two weighted connections are very much capable of approximating just about any functional mapping [15]. The MLP is mathematically represented by:

$$y_k = f_{outer}\left[\sum_{j=1}^{M} w_{kj}^{(2)} f_{inner}\left[\sum_{i=1}^{d} w_{ji}^{(1)} x_i + w_{j0}^{(1)}\right] + w_{k0}^{(2)}\right] \quad (1)$$

where $y_k$ represents the k-th output, $f_{outer}$ represents the output layer transfer function, $f_{inner}$ represents the input layer transfer function, $w$ represents the weights and biases, $^{(i)}$ represent the i-th layer.

*2) The radial basis function (RBF)*

In this class of neural networks, the activation of the hidden unit is determined by the distance between the input vector and the prototype vector [15]. The internal representation of the hidden units of the RBF network leads to a two stage training procedure. The first stage is concerned with the determination of the centre of the network using unsupervised methods. The second stage is concerned with the determination of the final-layer weights. The RBF networks provide a basis function (an interpolation function) which passes through each and every data point. A simple representation of the RBF network is depicted in figure 2. The RBF is mathematically represented by:

$$y_k(x) = \sum_{j=0}^{M} w_{kj} \phi_j(x) \quad (2)$$

where $y_k$ represents the k-th output, $w$ represents the weights and biases, and $\phi$ represents the activation functions of the output layer.

### D. The Theory of Support Vector Machines in Regression

Like ANNs, support vector machines (SVMs) can be used both for classification and regression problems. A support vector machine (SVM) is a classifier derived from statistical learning theory and was first introduced by Vapnik *et al* [16] in COLT-92. In regression problems, a non-linear function is learned by a linear learning machine in a kernel induced feature space, while the capacity of the system is controlled by a parameter that does not depend on the dimensionality of the space [13]. The process of employing SVMs in regression problems is referred to Support Vector Regression (SVR).

In SVR, the basic idea is to map the input space $x$ to the high dimensional feature space $\Phi(x)$ in a non-linear manner. This relationship is depicted in (3) where $b$ is the threshold.

$$f(x) = (w \cdot \Phi(x)) + b \quad (3)$$

Both $b$ and the constant $w$ are estimated by minimizing the sum of the empirical risk and a complexity term. In (4) below, the first term denotes the empirical risk, and the second term denotes the complexity term.

$$R_{reg}[f] = R_{emp}[f] + \lambda \|w\|^2 = C\sum_{i=1}^{Z}(f(x_i) - y_i) + \lambda \|w\|^2 \quad (4)$$

Where $Z$ denotes the size of the sample, $C(\cdot)$ is a cost function and $\lambda$ is the regularization constant.

## IV. STRUCTURED METHODOLOGY

This part of the paper describes the very structured methodology employed in order to get to the most optimum results of the study. The roadmap of this methodology is as depicted in fig. 1 below.

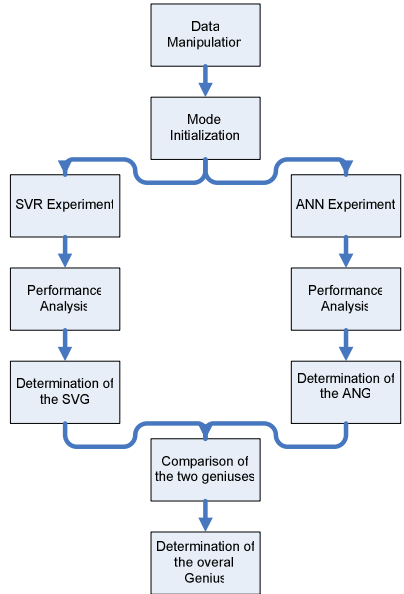

Fig. 1. The structured methodology adopted in this comparative study

The first stage of the methodology is to manipulate the data used in the study followed by the initialization of the model parameters. This stage is followed by two experiments that run in parallel, one for support vector regression and one for the artificial neural networks. A performance analysis is executed on both sides, and that is followed by the determination of the Support Vector Genius (SVG) and the Artificial Neural Genius (ANG). The SVG is the SVM model that outperforms all the other SVM models in the SVR experiment. The ANG is that ANN architecture that outperforms all the other models in the ANN experiment. The SVG and the ANG are thereafter compared in order to establish the overall Genius in the study.

## V. EXPERIMENTAL SETUP

### A) Data Manipulation

The data used in this study consists of the previous daily water demands and the annual estimated population size of the Gauteng Province. This data is manipulated in two forms, namely, normalization and division. The population figures depicted an increasing trend, as shown on table I, but the water demand figures are of arbitrary complexity as depicted in table II.

TABLE I
A SNAPSHOT OF THE ANNUAL POPULATION ESTIMATES

| Year | Mid-year Population Estimate |
|---|---|
| 1994 | 7 830 904 |
| 1995 | 7 992 219 |
| 1996 | 8 156 857 |
| 1997 | 8 324 886 |
| 1998 | 8 496 376 |

TABLE II
A SNAPSHOT OF THE DAILY WATER DEMAND FIGURES

| Date | Demand (Mega Liters) |
|---|---|
| 04-Jan-1997 | 1 849.95 |
| 05-Jan-1997 | 2 137.14 |
| 06-Jan-1997 | 1 982.94 |
| 07-Jan-1997 | 2 188.65 |
| 08-Jan-1997 | 2 254.14 |

*1) Data Normalization*

In order to simplify the task of the network, the data was scaled or normalized by making use of (5).

$$\tilde{x} = \frac{x - x_{MIN}}{x_{MAX} - x_{MIN}} \quad (5)$$

Where $\tilde{x}$ is the scaled data point, $x$ is the original data point, $x_{MIN}$ and $x_{MAX}$ are the minimum and maximum values in the data set, respectively. This is done in order to ensure that the minimum value in the data set is scaled to zero, and that the maximum value is scaled to one.

*2) Data Division*

The water figures obtained from Rand Water's database where comprised of 3 474 data points from the 4[th] of January 1997 to the 09[th] of July 2006. There was only one data point missing and this was on the 25[th] of March 1999. The effects of this missing data point were removed by discarding it from the database. Consequently the data bank remained with a sum of 3 473 data points. In order to employ the cross-validation technique, the data bank was divided into three interdependent data sets. These are the training set, the validation set and the testing set. The distribution and sum of these data sets is depicted in table III below.

TABLE III
THE DISTRIBUTION AND THE SUM OF DATA POINTS IN EACH DATA SET

| Data Set | Distribution | Total |
|---|---|---|
| Training Set | 294 × 5 | 1 470 |
| Validation Set | 201 × 5 | 1 005 |
| Testing Set | 199 × 5 | 995 |

### B) Model Initialization

This section deals with the issues of the number of model inputs. A short investigation had to be carried out and this was done from the ANN perspective. Initially the model is given a total of two inputs, followed by three, four, five and six inputs. A five input network reflects the least amount of training error and hence is adopted. The first four inputs are the previous water demand figures representing four consecutive days, and the fifth input is the annual population figure. A sample of the results from the model input development procedure is reflected in table IV below. This sample shows the results obtained from MLP architecture making use of the linear scaled conjugate gradient optimization algorithm.

TABLE IV
A SAMPLE OF THE RESULTS USED TO DECIDE ON THE NUMBER OF MODEL INPUTS

| Inputs | Training Error |
|---|---|
| Two | 1.585493 |
| Three | 1.552321 |
| Four | 1.525390 |
| Five | 1.538540 |
| Six | 1.539795 |

In order to facilitate fair comparison between ANNs and SVMs, the same number of model inputs was adopted for the SVR experiment.

VI. EXPERIMENTAL RESULTS AND ANALYSIS

*A) Object of the Performance Analysis*

The SVR experiment is carried out in parallel with the ANN experiment, and the performance of all the models is analyzed. The object of the analysis is to determine the genius model from each experiment. The genius model from the SVR experiment is referred to as Support Vector Genius (SVG) and the genius from the ANN experiment is referred to as the Artificial Neural Genius (ANG). The SVG and the ANG are then compared in order to determine the Overall Genius (OG). These two parallel experiments are simulated on a Pentium 4 computer with a frequency of 2.40GHz.

*B) Determination of the SVG*

In order to fine-tune the heuristics of the SVR models different kernel functions are tried and tested. Some of these kernels have additional arguments such as the degree, scale, offset, sigma (width) and maximum order of terms. The kernels that are available for use are the Anova, BSpline, exponential radial basis function (ERBF), Linear, Polynomial (Poly), radial basis function (RBF) and Spline. To determine the SVG, the different models are trained in a supervised manner and thereafter given the validation set to estimate the target of the validation set. The SVG is that model that has the least error and the most accuracy when estimating the target value of the validation set. This therefore implies that the two key performance parameters are the validation error and the accuracy. The other performance parameter taken into deliberation is the execution time, but does not carry much weight.

The validation error is computed using the traditional method of computing the percentage error. The accuracy of model can be evaluated in many ways. In this paper the accuracy is evaluated based on the practicality of the water demand figures. This is done by introducing a tolerance figure, $\tau$, with which the predicted value can be acceptable. This implies that the predicted value is regarded as accurate if it is equal to the actual value, plus or minus the tolerance figure. The sum of the accurate values is thereafter divided by the total number of points in the test set and multiplied by hundred to give the percentage of the accurate values in the validation set. This relationship is depicted in (6) below.

$$Acc = \frac{Cnt\ \{\forall\ |(prediction - actual)| \leq \tau\}}{Cnt\ \{\forall\ prediction\}} \times 100 \quad (6)$$

Where *Acc* represents the accuracy, *Cnt* denotes the count operation and $\tau$ is the acceptable tolerance.

According to the South African Department of Water Affairs and Forestry, the water services sector represents an overall demand of the order of 19% of the total water use [16]. This implies that 19% of the water used is consumed by the water supplier. This figure is therefore introduced as the tolerance value in the accuracy check. Nineteen percent of the average annual water demand (2700 Mega litres) is 500 Mega litres.

The results obtained from the SVR experiment are tabulated in table V below. The code 999 stands for 'NOT APPLICABLE'.

TABLE V
SUMMARY OF THE RESULTS OBTAINED FROM THE SVR EXPERIMENT

| Kernel | Degree | Scale | Offset | Sigma | Max Order | Error (%) | Accuracy (%) | Time (s) |
|---|---|---|---|---|---|---|---|---|
| Anova | 999 | 999 | 999 | 999 | 0 | 4.044 | 100 | 1294.1 |
| Anova | 999 | 999 | 999 | 999 | 1 | 4.044 | 100 | 1289.1 |
| Anova | 999 | 999 | 999 | 999 | 2 | 4.044 | 100 | 1269.3 |
| Anova | 999 | 999 | 999 | 999 | 3 | 4.044 | 100 | 1330.4 |
| BSpline | 0 | 999 | 999 | 999 | 999 | 4.95541 | 100 | 320.2 |
| BSpline | 1 | 999 | 999 | 999 | 999 | 11.9352 | 79 | 255.8 |
| ERBF | 999 | 999 | 999 | 1 | 999 | 4.28951 | 100 | 451.5 |
| ERBF | 999 | 999 | 999 | 2 | 999 | 4.29828 | 100 | 429.1 |
| ERBF | 999 | 999 | 999 | 3 | 999 | 4.29901 | 100 | 433.2 |
| Linear | 999 | 999 | 999 | 999 | 999 | 3.94003 | 100 | 3911.8 |
| Poly | 1 | 999 | 999 | 999 | 999 | 3.94016 | 100 | 2194.7 |
| Poly | 2 | 999 | 999 | 999 | 999 | 4.09741 | 100 | 11868.6 |
| Poly | 3 | 999 | 999 | 999 | 999 | 4.82196 | 100 | 18130.7 |
| RBF | 999 | 999 | 999 | 5 | 999 | 5.4237 | 98 | 1823.7 |
| RBF | 999 | 999 | 999 | 6 | 999 | 5.3654 | 99 | 507.3 |
| RBF | 999 | 999 | 999 | 7 | 999 | 5.22129 | 100 | 359.1 |
| Spline | 999 | 999 | 999 | 999 | 999 | 10.9939 | 83 | 467.7 |

It is evident from table V above that the model with the most optimum approximation is the one with a linear kernel function. This is due to the fact that it has 100% accuracy, and 3.94% validation error. It is therefore regarded as the Support Vector Genius (SVG).

*C) Determination of the ANG*

The ANN experiment has two architectures to investigate, and in turn, these architectures have many different activation functions. For the sake of simplicity, the experiments of the two architectures are separated and the results are compared.

*1) The MLP experiment and results*

The MLP network is trained using three different output unit activation functions and three different training algorithms. The activation functions are 'linear', 'logistic' and 'softmax'. The three different training algorithms are the scaled conjugate gradient (SCG), conjugate gradient (Conjgrad) and quasinewton (Quasinew) [17]. The softmax activation function gives a straight line approximation and hence its results are redundant. The experiment is therefore conducted with the other two activation functions and the three different optimization algorithms. The MLP ANN configurations are labelled as depicted in table VI. After the optimization of each of the network nomenclatures listed in table VI, the validation error analysis and accuracy check is executed using (6) and the results are shown in table VII.

TABLE VI
LABELING OF THE MLP ANNs ACCORDING TO THE ACTIVATION FUNCTION AND THE NUMBER OF HIDDEN UNITS

| ANN Label | Function | Units | Algorithm |
|---|---|---|---|
| AZ1 | Linear | 9 | SCG |
| AZ2 | Linear | 10 | SCG |
| AZ3 | Linear | 9 | Conjgrad |
| AZ4 | Linear | 10 | Conjgrad |
| AZ5 | Linear | 9 | Quasinew |
| AZ6 | Linear | 10 | Quasinew |
| AZ7 | Logistic | 9 | SCG |
| AZ8 | Logistic | 10 | SCG |
| AZ9 | Logistic | 9 | Conjgrad |
| AZ10 | Logistic | 10 | Conjgrad |
| AZ11 | Logistic | 9 | Quasinew |
| AZ12 | Logistic | 10 | Quasinew |

Both AZ2 and AZ11 have an accuracy of 99%. However AZ2 has a validation error that is less than that of AZ11. This therefore implies that the MLP ANN with the most suitable functional mapping is AZ2. AZ2 is a network with a linear output activation function, ten hidden units and the scaled conjugate gradient optimization algorithm.

TABLE VII
THE RESULTS OBTAINED FROM THE DIFFERENT MLP CONFIGURATIONS

| ANN | Error | Accuracy | Elapsed time |
|---|---|---|---|
| AZ1 | 23% | 38% | 76.954s |
| AZ2 | 6% | 99% | 81.360s |
| AZ3 | 32% | 5% | 184.109s |
| AZ4 | 10% | 87% | 156.828s |
| AZ5 | 63% | 0% | 73.594s |
| AZ6 | 35% | 7% | 20.875s |
| AZ7 | 15% | 73% | 96.703s |
| AZ8 | 6% | 97% | 20.281s |
| AZ9 | 9% | 93% | 90.781s |
| AZ10 | 18% | 59% | 154.984s |
| AZ11 | 7% | 99% | 76.515s |
| AZ12 | 9% | 96% | 146.968s |

*2) The RBF Experiment and Results*

The RBF network is trained in a manner that assesses the effects of three different activation functions. First, a network with Gaussian activations (Gaussian) is created and a two-stage training approach is used. It uses a small number of iterations of the Expectation-Maximization (EM) algorithm [17] to position the centres of the network and then the pseudo-inverse of the design matrix to find the second layer weights. The second layer has thin plate spline (TPS) activation functions and it makes use of the centres from the previous network to calculate the second layer weights. The third layer has $r^4 \log r$ (R4logr) activation functions. The combination of these activation functions and the number of the hidden units in the RBF network is labelled as in table VIII. Similarly, after the optimization of each of the RFB network the error analysis (validation error) and accuracy check is executed using (4) and (5) respectively.

TABLE VIII
LABELLING OF THE RBF ANNS ACCORDING TO THE ACTIVATION FUNCTIONS AND THE NUMBER OF HIDDEN UNITS

| ANN Label | Function | Units |
|---|---|---|
| AX1 | Gaussian | 9 |
| AX2 | Gaussian | 10 |
| AX3 | TPS | 9 |
| AX4 | TPS | 10 |
| AX5 | R4logr | 9 |
| AX6 | R4logr | 10 |

Table IX shows ANN configurations with 100% accuracy. These are AX3, AX4, AX5 and AX6. In order to select the most optimum one, the validation error is observed to select the smallest. Both AX4 and AX6 have the same smallest validation error. In order to select the most optimum one, the error obtained during training is observed.

**AX4 Training Error = 2.4651%**
**AX6 Training Error = 2.4272%**

TABLE IX
THE RESULTS OBTAINED FROM THE RBF VALIDATION FOR THE DIFFERENT ACTIVATION FUNCTIONS

| ANN | Error | Accuracy | Elapsed Time |
|---|---|---|---|
| AX1 | 28% | 37% | 12.969s |
| AX2 | 15% | 71% | 9.671s |
| AX3 | 3.7% | 100% | 12.969s |
| AX4 | 3.6% | 100% | 9.671s |
| AX5 | 4.2% | 100% | 12.969s |
| AX6 | 3.6% | 100% | 9.671s |

This is a small difference but AX6 has the smallest training error and hence the most optimum functional mapping. This therefore implies that the best RBF is AX6. When comparing AZ2 and AX6 it is apparent that AX6 provides the most optimum approximation of the target values of the validation set. This is due to the fact that it has 100% accuracy and a validation error of 3.6%. As a result, it is regarded as the Artificial Neural Genius (ANG).

## VII. DISCUSSION OF RESULTS

The object of this section is to compare the SVG and ANG in order to determine the OG. This is done by employing the third data set which is the testing set. This is done in order to determine the generalisation ability of these two geniuses.

| | |
|---|---|
| **SVG Error** | **5.46519%** |
| **SVG Accuracy** | **100%** |
| **ANG Error** | **2.95995%** |
| **ANG Accuracy** | **100%** |

It is apparent from this analysis that the ANG has better generalization ability than the SVG. This therefore implies that artificial neural networks are better approximation tools for this particular study. The functional mapping of the ANG is plotted in the fig 2 below.

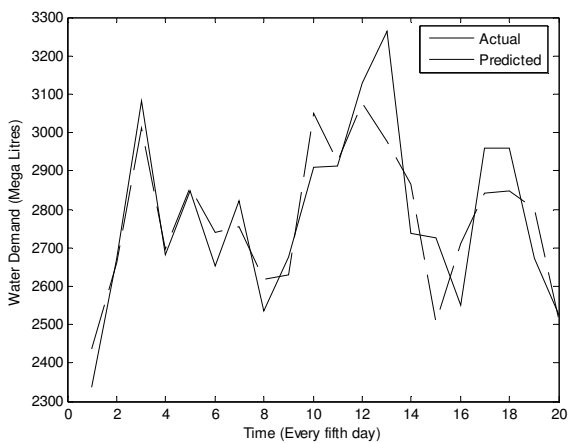

Fig. 2. The functional mapping of the ANG

A further development would be to even further scrutinize the generalisation ability of this ANG. This can be done by determining the common trends in the water demand data. A model will be developed for each trend, and the theory of Hidden Markov Models [18] will be employed to determine whether the predicted value belongs to that particular date or season by looking at the model of that particular season.

## VIII CONCLUSIONS

Two machine learning techniques have been investigated in this study. These are the artificial neural networks (ANNs) and the support vector machines (SVMs). An approach adopted was to conduct two parallel experiments, one for the ANNs and one for the SVMs. The ANN experiment encapsulated two architectures, the multi-layer perceptron (MLP) and the radial basis function (RBF). The results from the two architectures were compared to come up with the Artificial Neural Genius (ANG). The SVM experiment was comprised of many models with different kernel functions and some of these kernel functions had additional arguments such as the degree and the scale. These models were compared against each other in order to determine the Support Vector Genius (SVG). The performance criteria used to determine the geniuses from each experiment were the validation error and the accuracy in their approximation of the target values of the validation data set. The two geniuses were then compared against each other in order to determine the overall genius (OG). The performance parameter used to determine the OG is the generalisation ability of each genius. The ANG has proved to outperform the SVG.

## ACKNOWLEDGEMENT
The authors hereby thank Rand Water's Thomas Phetla for taking his time to make the data available. The financial support of the South African National Research Foundation is hereby acknowledged.